\documentclass[10pt, a4paper]{article}

\usepackage[final]{lrec2026} 
\usepackage{booktabs}
\usepackage{tabularx}
\usepackage{multirow}
\usepackage{array}
\usepackage{xcolor}
\usepackage{alltt}
\usepackage{listings}
\usepackage{natbib}
\usepackage{hyperref}

\lstdefinestyle{umr}{
  basicstyle=\ttfamily\small,
  columns=fullflexible,
  keepspaces=true,
  showstringspaces=false,
  escapeinside={(*@}{@*)}
}

\usepackage{amsmath}

\usepackage{xurl}

\title{Enhancing Structured Meaning Representations with Aspect Classification}

\name{Claire Benét Post$^*$, Paul Bontempo$^*$, August Milliken$^*$, \\Alvin Po-Chun Chen, Nicholas Derby, Saksham Khatwani, \\ Sumeyye Nabieva, Karthik Sairam, and Alexis Palmer}

\address{University of Colorado, Boulder}

\abstract{
To fully capture the meaning of a sentence, semantic representations should encode \textit{aspect}, which describes the internal temporal structure of events. In graph-based meaning representation frameworks such as Uniform Meaning Representations (UMR), aspect lets one know how events unfold over time, including distinctions such as states, activities, and completed events. Despite its importance, aspect remains sparsely annotated across semantic meaning representation frameworks. This has, in turn, hindered not only current manual annotation, but also the development of automatic systems capable of predicting aspectual information. In this paper, we introduce a new dataset of English sentences annotated with UMR aspect labels over Abstract Meaning Representation (AMR) graphs that lack the feature. We describe the annotation scheme and guidelines used to label eventive predicates according to the UMR aspect lattice, as well as the annotation pipeline used to ensure consistency and quality across annotators through a multi-step adjudication process. To demonstrate the utility of our dataset for future automation, we present baseline experiments using three modeling approaches. Our results establish initial benchmarks for automatic UMR aspect prediction and provide a foundation for integrating aspect into semantic meaning representations more broadly. \\ \newline  
}

\begin{document}

\maketitleabstract
\footnotetext[1]{$^*$These authors contributed equally to this work.}

\section{Introduction}\label{sec:intro} 




Semantic representations frequently center around capturing components of meaning related to the core \emph{events} conveyed by individual natural language utterances. 
Nearly all meaning representation (MR) formats express the core predicates associated with those events, along with any arguments to those predicates. 
MRs differ quite substantially, though, when it comes to the expression of additional event information, such as tense, modality, aspect, or information structure. 
Languages also differ substantially in the degree to which they grammaticalize (or require the expression of) the same event-related information.

Aspect is a core component of Uniform Meaning Representation (UMR), a graph-based semantic framework designed to represent meaning in a cross-linguistically applicable and computationally tractable way. 
Unlike tense, which encodes \textit{when} an event occurs, aspect captures the \textit{how}: the internal temporal structure, duration, and completedness of events~\citep{comrie1976aspect}, \citep{croft2012verbs}, \citep{donatelli2018annotation}. 
It allows a semantic system to distinguish between, for example, habitual, ongoing processes, or completed achievements, enabling a more nuanced interpretation of event semantics.


\begin{figure}[t]
\centering
\begin{lstlisting}[style=umr]
She (*@\textcolor{magenta}{is still writing}@*) her paper.
(w/ (*@\textcolor{magenta}{write-01}@*)
    :ARG0 (p/ person
        :ref-person 3rd
        :ref-number Singular)
    :ARG1 (p2/ paper
        :poss p
        :ref-number Singular)
    :mod (s/ (*@\textcolor{magenta}{still}@*))
    :aspect (*@\textcolor{magenta}{Activity}@*)
    :modstr FullAff)
\end{lstlisting}
\caption{Example UMR graph with the eventive highlighted with an \textit{Activity} aspectual marker.}
\label{fig:activity}
\end{figure}




In UMR, aspect is applied to all \textit{eventive} elements (also known as \textit{eventualities}) in a sentence. The central eventuality introduced by an utterance
is typically the concept aligned with the main finite verb, as seen in Fig.~\ref{fig:activity}. Eventualities in UMR 
refer to the full predication, encompassing the verb and its arguments~\citep{donatelli2019tense, kingsbury2003propbank}. UMR defines a particular inventory of aspectual categories, aligned with other well-established event typologies, including (among others) states, activities, accomplishments, achievements, and processes~\citep{bach1986algebra}. In UMR the aspectual categories are organized into a lattice that supports both coarse- and fine-grained aspectual distinctions. Unlike surface-level grammatical cues, like those in auxiliaries or verb morphology, UMR aspect is a semantic feature.
It abstracts away from morphosyntactic form to represent covert event structure and is intended to 
generalize across typologically diverse languages~\citep{van2021designing}.

Annotating aspect is no simple feat. Theoretical debates span decades, including disagreements about the universality of aspectual categories, the granularity of classifications, and their interaction with tense and modality~\citep{reichenbach1947elements, vendler1957verbs, comrie1976aspect, langacker2011remarks, dowty1986effects, hinrichs1986temporal, moens1988temporal, klein2013time, chang2022dynamic, partee2011nominal, croft2012verbs}.
While there are a number of corpora with aspect annotations, and several computational models \citep{friedrich-etal-2023-kind}, there is no unified approach to annotation or modeling~\citep[among others:][]{pustejovsky2003timeml, derczynski2017automatically, pustejovsky2017designing, friedrich2014automatic, friedrich2016situation, mostafazadeh2016caters, laparra2018characters, o2016richer, gantt-etal-2022-decomposing}.

From a typological perspective, some languages encode aspect more saliently than others, further complicating annotation for multilingual or cross-linguistic frameworks. For example, American Sign Language and Mandarin Chinese prioritize aspectual distinctions over tense~\citep{li1989mandarin, mcdonald1982aspects}, while Hindi includes a dedicated aspect morpheme separate from tense or mood~\citep{van1975aspect}. In contrast, many Indo-European languages conflate aspect and tense morphologically, often obscuring the underlying semantic distinctions.


Given these complexities, manual aspect annotation is time-consuming, error-prone, and highly sensitive to annotator interpretation. Yet its inclusion in UMR is a cornerstone to achieving a more expressive, cross-linguistic meaning representation system. UMR builds on earlier formalisms such as Abstract Meaning Representation (AMR)~\citep{banarescu2013abstract}, where aspect was initially introduced to support event-based reasoning but was never fully adopted into standard annotation guidelines. \citet{donatelli2018annotation, donatelli2019tense} formalize aspect in AMR, laying out annotation principles and aligning event types with lexical frames.


Despite its importance for accurately representing event semantics, aspect remains under-annotated in existing UMR resources. This scarcity limits both the scale and consistency of manual UMR annotation and hinders development of automatic parsers capable of reliably predicting aspect.

To address this bottleneck, we present a new dataset of English sentences manually annotated with UMR aspect labels. 
Annotators are provided with AMR-derived graphs for each sentence and asked to label eventualities according to the UMR aspect lattice (see~\ref{sub:dataSource} for corpus details). 
AMR graphs can be converted to UMR graphs with a combination of automated processing and manual intervention, and \citet{bonn-etal-2023-mapping} outline the correspondences and differences between the two formalisms. 
By adding aspect annotations, the new dataset provides a bridge to complete UMR graphs.

Our annotation pipeline combines structured annotator training with multiple rounds of independent annotation, group adjudication sessions, and expert consultation to improve the guidelines and resolve difficult annotation decisions, resulting in a high-quality resource intended to support future UMR modeling efforts. To validate dataset quality,
we additionally report baseline experiments spanning three modeling families: (1) a rule-based approach, (2) an embedding-based classifier, and (3) a large language model (LLM) prompting approach.

We target two complementary objectives:

\begin{itemize}
    \item \textbf{Task 1: Data annotation.} We construct a new gold-standard dataset of carefully validated sentences labeled according to the UMR aspect annotation scheme.
    \item \textbf{Task 2: Baseline modeling.} We present standard data splits and initial performance benchmarks for automated UMR aspect labeling.
\end{itemize}

To support both tasks, we 
perform a large-scale annotation project, resulting in a new dataset of $1,473$ manually annotated aspect labels for the eventive predicates of AMR graphs. 
This is the first dataset designed to support supervised learning approaches for UMR aspect annotation.
\section{Related Work}\label{sec:related} 

\begin{figure*}[t]
  \includegraphics[width=\linewidth]{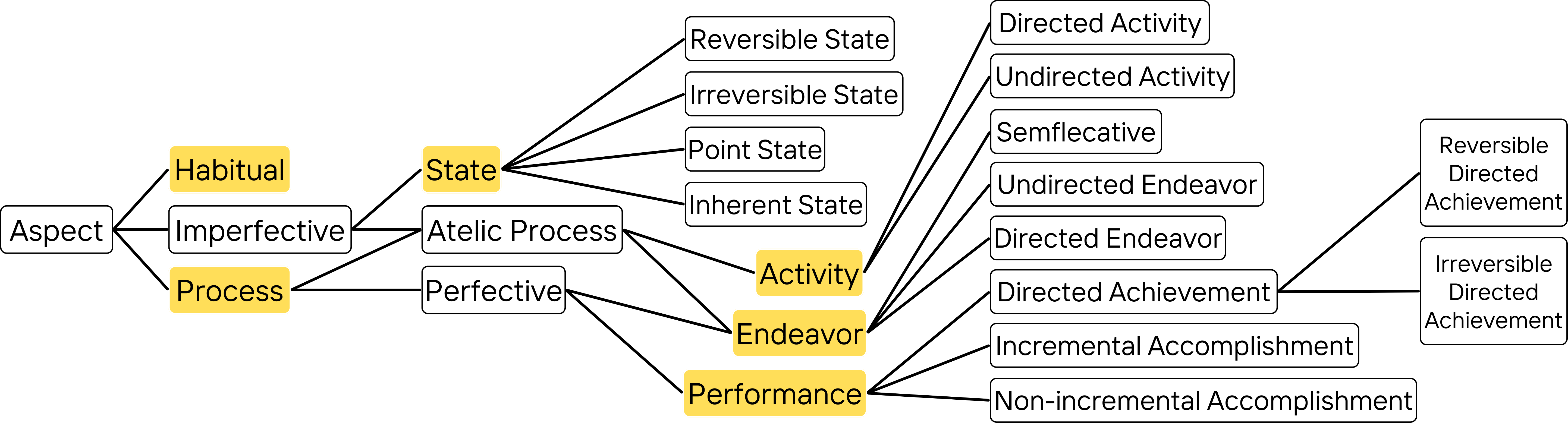}
  \caption{UMR aspect lattice with aspectual values utilized in our English annotation highlighted in yellow.}
  \label{fig:aspect-english}
\end{figure*}





The semantics of aspect has been a long-standing topic of debate in linguistic theory. Seminal works by \citet{reichenbach1947elements}, \citet{vendler1957verbs}, and \citet{comrie1976aspect} lay the foundation for distinguishing between types of eventualities--states, achievements, activities, accomplishments--based on their temporal and structural properties. \citet{dowty1986effects} and \citet{langacker2011remarks} further explore the interaction between aspect, argument structure, and lexical semantics. These formalisms inform how events are modeled in UMR today.

Later developments such as Hinrichs’ interval-based models \shortcite{hinrichs1986temporal}, Moens and Steedman's narrative structure theory \shortcite{moens1988temporal}, and Klein’s temporal logic \shortcite{klein2013time} introduce more formal ways to encode event structure and its temporal entailments. These insights highlight the need for meaning representation frameworks like UMR to go beyond grammatical tense and directly encode aspectual distinctions based on semantic content.


\paragraph{Aspect annotation.} Aspect has been incorporated into previous semantic annotation and event modeling efforts, particularly in temporal information extraction. TimeML~\citep{pustejovsky2003timeml} and its follow-up projects such as the TempEval competitions~\citep{derczynski2017automatically} include annotation for aspect, though typically via shallow textual cues. There are several datasets developed with robust manual aspect annotation that consider sentential context and event structure, such as DIASPORA~\citep{kober-etal-2020-aspectuality} and the Universal Decompositional Semantics dataset~\cite{Gantt-UDS-dataset-2021}; however, these datasets employ coarse-grained aspect classes, rather than the more extensive and typologically-applicable lattice provided by 
the UMR framework. More recent work seeks to automate aspect classification using linguistic features~\citep{friedrich2014automatic}, discourse roles~\citep{friedrich2016situation}, and LSTM-based models that integrate context~\citep{mostafazadeh2016caters, laparra2018characters}.

While effective to some degree, these systems often operate over flat text or shallow syntactic representations. They do not handle the rich predicate-argument structures or graph-based semantics found in UMR and AMR. Moreover, they treat aspect as a kind of downstream feature, rather than an integral part of event structure representation.

Efforts to include aspect in AMR were initiated by \citet{donatelli2018annotation}, but aspect is not part of AMR's core schema. On the other hand, UMR incorporates aspect explicitly into its annotation guidelines, enabling more structured reasoning about events across languages. 

\paragraph{Aspect Annotation in UMR.} 






Due to the small amount of available UMR data, prior work has focused primarily on methods for generating UMR graphs without supervised training. \citet{chun-xue-2024-pipeline} propose a multi-step strategy for converting AMR graphs into UMR graphs by using a variety of existing automation tools, such as by using a modal dependency parser. Similarly, \citet{sun-etal-2024-chinese} experiment with few-shot and Think-Aloud prompting on LLMs to generate Chinese UMR graphs without AMR data as input. \textsc{AutoAspect}, which directly targets UMR aspect, proposes a rule-based approach specifically for classifying UMR aspects in English UMR graphs~\citep{chen-etal-2021-autoaspect}.

While these approaches produce some positive results, we find in our investigations that training on a larger dataset is necessary for further improvement. Moreover, significant refinements have been made to the UMR dataset since these resources were published, as seen in~\citep{bonn2024building}.

\section{Annotation Scheme}\label{sec:annotation}


\subsection{UMR Aspect Lattice}




This paper is concerned with the aspect annotation of English sentences, as highlighted in \autoref{fig:aspect-english}. 
The aspectual categories chosen for English annotation include a set of base-level distinctions--\textit{State}, \textit{Performance}, \textit{Endeavor}, \textit{Activity}, and \textit{Habitual}, and \textit{Habitual}--as well as a more coarse-grained value for event nominals and other underspecified events, \textit{Process}. 


UMR organizes aspectual categories within an aspect lattice, a hierarchical structure that captures relationships between coarse and fine-grained labels. This design allows annotations to represent general distinctions while remaining compatible with more specific readings when additional linguistic information is available. 

This structure is particularly useful for cross-linguistic semantic annotation. English often relies on relatively coarse-grained aspectual distinctions, while other languages encode finer aspectual contrasts directly in their grammar. The lattice enables UMR to support consistent representations across typologically diverse languages. This was a major motivation for our work, and we hope in the future to expand aspect annotation to more languages.

\subsection{Aspect Types}
\paragraph{\textit{State.}} 
This value corresponds to stative events, indicating that no change occurs during the event, as prescribed by \citep{vendler1967linguistics}. It includes predicate nominals, predicate locations, and thetic (presentational) possession. 
\\ \\
\noindent [1] \emph{The cat loves milk}.
\begin{small}
\begin{alltt}
(l/ \textcolor{magenta}{love-01}
    :ARG0 (c/ cat
        :ref-number Singular)
    :ARG1 (m/ milk)
    :aspect \textcolor{magenta}{State}
    :modstr FullAff)
\end{alltt}
\end{small}
\noindent
More specifically, in English, the \textit{State} value encompasses modal verbs (e.g., "The cat \textbf{needs} to eat.") and events under the scope of ability modals (e.g., "The cat is \textbf{able} to eat."). UMR classifies \textit{inactive actions}, as defined by \citep{croft2012verbs}, as stative. This includes posture verbs (e.g., "The cat \textbf{hangs} on the windowsill."), perception verbs (e.g., "The cat \textbf{sees} milk."), mental activities (e.g., "The cat \textbf{thinks} about jazz."), verbs of operation (e.g., "The cat is \textbf{working} on catching mice."). The \textit{State} value, in English, is also an umbrella that covers inherent states (e.g., "The cat \textbf{is} black."), reversible states (e.g., "The cat is \textbf{hungry}."), irreversible states (e.g., "The glass \textbf{is shattered}."), and point states (e.g., "When it \textbf{is} 12:30pm, feed the cat.").

\paragraph{\textit{Performance.}} This category covers events that reach a result state, such as achievements that have some instantaneous binary change, accomplishments where there is a run-up process before the change, or when the event reaches a result state that has a natural endpoint. 
\\ \\
\noindent [2] \emph{The cat walked along the fence in 2 minutes.}
\begin{small}
\begin{alltt}
(w/ \textcolor{magenta}{walk-01}
    :ARG0 (c/ cat
        :ref-number Singular)
    :ARG2 (a/ along
        :op1 (f/ fence))
    :duration (t/ temporal-quantity
        :unit (m/ minute)
        :quant 2)
    :aspect \textcolor{magenta}{Performance}
    :modstr FullAff) 
\end{alltt}
\end{small}
\noindent
For instance, completive markers (e.g., "The cat finished \textbf{climbing} \textit{up the tree}.") and container adverbials (e.g., "The cats \textbf{scampered} along the fence \textit{in 10 seconds}.") are both indicators that an event has reached a distinct result state.




\paragraph{\textit{Endeavor.}} The \textit{Endeavor} value is often mistaken for \textit{Performance} and vice versa. 
\textit{Endeavor} is used for processes that end within the time window in question, but do \textit{not} reach a particular result state: e.g., compare graph [2] to graph [3] below.
\\ \\
\noindent [3] \emph{The cat walked along the fence.}
\begin{small}
\begin{alltt}
(w/ \textcolor{magenta}{walk-01}
    :ARG0 (c/ cat
        :ref-number Singular)
    :ARG2 (a/ along
        :op1 (f/ fence))
    :aspect \textcolor{magenta}{Endeavor}
    :modstr FullAff)
\end{alltt}
\end{small}
\noindent
Generally, \textit{Endeavor} requires an explicit aspectual marking in English to be considered an \textit{Endeavor}. Terminative aspectual markers, like "stop" in English, and durative adverbials (e.g., "The cat \textbf{ate} kibble \textit{for thirty seconds}.") are both strong indicators for \textit{Endeavor} as they took place for a period of time then ended, likely without completion.

\paragraph{\textit{Activity.}} The \textit{Activity} aspect covers processes that do not start or end during the time window in question. They can be ongoing with respect to present or past time (e.g., “The cat \textbf{was playing} the piano.”) For an example graph, see \autoref{fig:activity}.

Identifying \textit{Activity} is difficult because it is largely dependent upon context, document creation time, and real world knowledge. However, there are some grammatical clues that can help one put together the puzzle. For example, if the event is in the present progressive (e.g., "The cat \textbf{is playing} the piano."), it is typically annotated with \textit{Activity}. Inceptive and continuative aspectual markers may also imply that an event has not ended (e.g., "The cat \textbf{started playing} the piano." and "The cat \textbf{kept on playing} the piano."). 

\paragraph{\textit{Habitual.}} The \textit{Habitual} aspectual sense is usually straightforward to identify. It covers things that happen repeatedly or regularly. One hallmark, in English, are phrases such as “used to” and “always” modifying the verb. Note that \textit{Habitual} does not require these cue phrases--as in the next example. 
\\ \\
\noindent [4] \emph{The cat eats kibble.}
\begin{small}
\begin{alltt}
(w/ \textcolor{magenta}{eat-01}
    :ARG0 (c/ cat
        :ref-number Singular)
    :ARG1 (k/ kibble)
    :aspect \textcolor{magenta}{Habitual}
    :modstr FullAff)
\end{alltt}
\end{small}
\noindent
In English, \textit{Habitual}  is often expressed through simple present construction, while the past is often done with "used to" (e.g., "The cat \textbf{used to eat} kibble.").

\paragraph{\textit{Process.}} This aspect is the most coarse-grained label available for English. It describes an ongoing event where the beginning or end is uncertain or unspecified. The most common use of \textit{Process} is as the default label for event nominalizations (e.g., "The cat denied \textbf{wrongdoing}.").
\\ \\
\noindent [5] \emph{After the game, the cat slept.}
\begin{small}
\begin{alltt}
(s/ sleep-01
    :ARG0 (c/ cat
        :ref-number Singular)
    :temporal (a/ after
        :op1 (g/ \textcolor{magenta}{game}
            :aspect \textcolor{magenta}{Process}))
    :aspect State
    :modstr FullAff)
\end{alltt}
\end{small}
\noindent
Graph [5] shows another category of events typically annotated with \textit{Process} in UMR. Here, the \textbf{game} event is packaged in a referring expression.
We take a similar approach for underived nominals, nominalizations, and gerunds.

\subsection{Comparison to Other Aspect Annotation Schemata}





Aspect annotation has been widely studied in linguistics, and UMR represents a recent schema that builds on prior theoretical work. In particular, UMR follows approaches such as \citet{croft2012verbs}, which emphasize that aspectual interpretation depends on multiple factors and that a single event may admit multiple plausible aspectual readings depending on context. This perspective informed our adjudication process when resolving difficult cases.


Because our task is motivated by downstream NLP and machine learning applications, our annotation process follows principles outlined by \citet{pustejovsky2017designing}. We prioritize consistency in label assignment in order to maximize the learnable signal in the dataset, even when this means limiting the number of annotated examples. Nevertheless, as shown in \autoref{fig:aspect-english}, the selected aspect labels differ in granularity, which introduces variation in specificity across the dataset.


Prior work such as the DIASPORA dataset \citep{kober-etal-2020-aspectuality} also explores aspect annotation but employs a more coarse-grained 3-label schema (\textit{state}, \textit{telic}, \textit{atelic}) to reduce label overlap and maintain uniform granularity. Our dataset instead adopts the richer UMR aspect inventory while remaining compatible with prior work. To illustrate this compatibility, we developed a mapping between our schema and the DIASPORA labels and automatically applied it to a subset of our data, manually evaluating the resulting label assignments. We found that the two schemata are broadly compatible, excepting the case of event nominals, which take \textit{process} labels in UMR but are without a consistent equivalent in DIASPORA, requiring manual adjudication rather than automated label mapping.

\section{Building the Corpus}\label{sec:data}

\subsection{Data}\label{sub:dataSource}



Our dataset is sourced from the UMR 2.0 Dataset \cite{umr-2.0-data-release-2025} which contains roughly 30k UMR graphs in different stages of conversion from AMR graphs \citep{Knight2020-rg,bonn-etal-2020-spatial}. 
Some UMR 2.0 graphs have aspect annotations from previous work; we annotate graphs that do not yet have aspect labels. To ensure broad coverage for training and evaluation, we select four corpora from the dataset to annotate:
\begin{enumerate}
    \item The Little Prince corpus, a set of sentences from the English translation of The Little Prince by Antoine de Saint-Exupéry.
    \item The Minecraft corpus, a set of dialogues and corresponding grounding data from a collaborative structure-building task in Minecraft~\citep{narayan-chen-etal-2019-collaborative}.
    \item The BOLT DF corpus, which contains English-language forum posts crawled as part of the DARPA BOLT project.
    \item The Weblog corpus, comprised of weblog and online news articles.
\end{enumerate}

A detailed summary of aspect label statistics by corpora for the existing UMR dataset can be found in Appendix A under \autoref{tab:gold_label_stats}.


\subsection{Annotation}



Annotation proceeded in two phases: in \textsc{Phase 1}, a team of 8 annotators worked in pairs to label each event from the partially-converted AMR graphs in bulk. In \textsc{Phase 2}, a smaller group focused on adjudicating decision ties with expert consultation, while ensuring consistency with previous annotations.  

\subsubsection{Phase 1}

Given the complexity of aspect annotation and its theoretical underpinnings, we determined that all members of the team should first develop a strong understanding of the UMR aspect schema, before proceeding to the bulk annotation of \textsc{Phase~1}. Our goals were to ensure that each annotator contributed quality data and that rules were applied consistently between annotators. To this end, we conducted 8 weekly training sessions throughout \textsc{Phase 1}. 


Annotation guidelines and training materials were built from existing UMR resources and developed into task-specific training materials.\footnote{See \autoref{tab:data_stats} in Appendix B for more details.}
Each week, team members presented on different topics from these materials and discussed example annotations as a group to clarify issues.
\footnote{Annotation education material: \url{https://drive.google.com/drive/folders/1_ou3WW4UV7gQHtglMTEbu4TlTbO5xdHt?usp=share_link}}

We conducted an initial practice task in which each team member annotated up to 50 predicates from the Pear Story corpus~\citep{umrdatarelease}. This dataset was selected for its short, visually grounded sentences, which aided learning and facilitated discussion. We reviewed inter-annotator agreement and recurring errors on the practice task before proceeding to bulk annotation. 


Label-wise accuracy showed that some categories such as \textit{State} and \textit{Performance} were more reliably identified, while minority classes like \textit{Endeavor} and \textit{Habitual} were less consistent. These findings guided a focused error correction in which we reviewed common sources of confusion, such as distinctions between \textit{State} vs. \textit{Performance} and \textit{Performance} vs. \textit{Endeavor}. 

Following the practice task with the Pear Story corpus, we moved into full-scale corpus annotation. Each sub-corpus was assigned to two annotators for independent labeling, resulting in two first-pass labels per event per corpus. Each numbered predicate within the AMR graphs was annotated with one of the six UMR aspect labels or marked with \textit{NONE} if the predicate was deemed non-eventive. The \textit{NONE} label was frequently used for adjectival or adverbial concepts, which often receive automatic FrameNet mappings in AMR (in order to provide conceptual context) but do not participate in eventualities. \autoref{tab:iaa} shows the distribution of aspect labels for the completed dataset. The first-pass bulk annotation lasted approximately 6 weeks.

\subsubsection{Phase 2}


Following \textsc{Phase 1}, all events with conflicting aspect labels were routed to a tie-breaking process. Each sentence and its annotations were reviewed by a third annotator   
to make a final determination. 
If the adjudicating annotator disagreed with both original labels, the sentence went to an additional adjudication step, up to a maximum of 5 total annotation rounds.
All intermediate labels are preserved and ranked in our final dataset, along with the final adjudicated labels.\footnote{In Appendix B, \autoref{tab:annotation_counts} shows the error rates for each label and the most commonly mistaken aspect labels.}

In the next step, a team of two annotators reviewed all data for consistency. Together, the tie-breaking and consistency adjudication lasted about 8 weeks.
This review process ensured each adjudicated aspect label:
(i) follows our annotation guidelines, and (ii) is consistent with other instances in the dataset. To confirm consistency of a given label, we compared against both sentences with the same event, and events with the same label.


The duration and complexity of this annotation process indicates the corresponding complexity of aspect itself; even with months of discussion, some instances in the dataset remained ambiguous. 
For particularly complex disagreements—such as differentiating \textit{Endeavor} from \textit{Performance}—we consulted directly with external experts to align the annotations with their interpretations.


\begin{table}[t]                                                                                             
\centering                                                                                                   
\small                                                                                                       
\begin{tabular}{lrr}                                      
\toprule
\multicolumn{3}{c}{\textbf{Per-class}} \\
\textbf{Label} & \textbf{$K\alpha*$} & \textbf{Count} \\
\midrule
None        & 0.735&  514 \\
State       & 0.706&  385 \\
Performance & 0.716&  360 \\
Process     & 0.547&  100 \\
Habitual    &  0.704&  56 \\
Activity    &  0.502&  40 \\
Endeavor    &  0.559&  18 \\
\midrule
\textbf{Total} & &  \textbf{1{,}473} \\
\midrule
\multicolumn{3}{c}{\textbf{Overall}} \\
\midrule
Percent agreement                             & & 74.1\%  \\
Cohen's $\kappa$                              & & 0.656   \\
Krippendorff's $\alpha$                       & & 0.688   \\
Fleiss' $\kappa$ $^\dagger$                   & & 0.146   \\
\bottomrule
\end{tabular}
\caption{Aspect label distribution and inter-annotator agreement metrics.
*Per-class Krippendorff's alpha computed as one-vs-rest.
$^\dagger$Fleiss' $\kappa$ restricted to the 333 events that received first-pass tie-breaks.}
\label{tab:iaa}
\end{table}
\begin{table*}[htbp]
\centering
\begin{tabularx}{\textwidth}{Xccc}
\toprule
Sentence & Event & Main label & 2nd label \\
\midrule
(a) <Architect> now above that place red blocks on the grid. & Place & Performance & Process \label{row:place_data}\\
(b) "Clap your hands, one against the other," the conceited man now directed him. & Clap & Endeavor & Performance \label{row:clap_data}\\ 
(c) Greed is when you are wealthy and lobby your representatives for special tax breaks because you are over 60 years of age. & Age & State & N/A \label{row:age_data}\\
\bottomrule
\end{tabularx}
\caption{Selected examples with ambiguous aspect; some annotated with secondary labels. 
}
\label{tab:difficult_examples}
\end{table*}


\subsection{Inter-Annotator Agreement}

We report agreement metrics across the various phases of our annotation and adjudication process, compared against our finalized gold-level labels in \autoref{tab:iaa}. Overall agreement represents the percentage of events where both first-pass annotators provided the same label for an event, though this does not necessarily mean that such events were exempt from adjudication later. Cohen's $\kappa$ is computed over all first-pass annotations, as another metric of agreement during \textsc{Phase 1}. Krippendorff's $\alpha$ compares all 5 rounds of tie-breaks (ignoring missing values where tie-breaking wasn't needed), meaning it best captures the agreement of all annotators as a group across both phases. We report Fleiss' $\kappa$, which allows for 3 or more annotators, for the events which moved from \textsc{Phase 1} (first-pass annotation) to \textsc{Phase 2} (adjudication) due to a label discrepancy between the two first-pass annotators and required a tie-breaker.

\section{Annotation Challenges}\label{sec:discussion}



In \autoref{tab:difficult_examples} we present examples representative of particularly ambiguous annotation cases. Many of these issues stem from the limited contextual information available to annotators, as sentences were presented individually rather than as part of complete documents. In naturally occurring discourse, surrounding context typically resolves such ambiguities, and aspectual interpretation is no exception. 

For instance, in \autoref{tab:difficult_examples} example (a), the event suggests a \textit{Process} label in the sentence context, since the speaker specifies no explicit number of blocks, and the event lacks an inherent endpoint. However, the wider discourse context (blocksworld video game) includes a particular number of red blocks available to the players, and a finite grid on which to place them, which instead motivates a \textit{Performance} label. 

Similarly, in example (b), contextual cues could determine whether an event is an \textit{Endeavor} or a \textit{Performance}, depending on whether we understand "clap" as a process without change of state (hands end up as they started), or a complete event that reaches a natural conclusion (hands start apart and end together in a single motion). Without surrounding document context, we cannot say for sure which type of clap this sentence describes. 

To account for multiple plausible interpretations, our adjudication schema allows for the identification of a secondary label that reflects a reasonable alternative reading, even when a primary label is selected based on the most likely interpretation. For events without such ambiguity such as example (c), we provide only one adjudicated label. 

\section{Automatic Annotation}\label{sec:methods} 

One goal of this annotation effort is to build training and evaluation data for automatic aspect classification.
Toward that end, we establish standard data splits and evaluate the performance of several baseline models.



\subsection{Data splits}

We establish standardized splits of our final dataset in a 70/15/15 train/val/test ratio, as illustrated in \autoref{tab:dataset-splits}, found in Appendix C. We split the data on sentence boundaries, rather than by event, to ensure that no contextual information is shared across splits, since a single sentence may include multiple events, and sentential context seen in training for one event could unfairly inform test performance on another. We stratified each split to ensure consistent and balanced class distribution.
To accomplish this, we first identify a dominant aspect class for each sentence, by counting the most frequent label across all events per sentence. For sentences with just one event or whose events all have different labels, we consider the first event's label to be dominant. We perform an initial 70/30 split of the sentences, keeping the label distribution of both consistent with those of the overall dataset. We do the same for the secondary division on the 30\% split, producing comparable validation and test splits. 



\begin{table*}[t]                                         
\centering
\small
\begin{tabular}{llrrrrr}
\toprule
\textbf{Type} & \textbf{Model} & \textbf{Acc.} & \textbf{Macro F1} & \textbf{Wtd.\ F1} & \textbf{Precision} &
\textbf{Recall} \\
\midrule
\multirow{4}{*}{LLM} & LLaMA-3.1-8B-Instruct (zero-shot) & 0.31 & 0.19 & 0.27 & 0.29 & 0.24 \\
& LLaMA-3.1-8B-Instruct (3-shot) & 0.25 & 0.16 & 0.22 & 0.32 & 0.21 \\
& GPT-5mini (zero-shot) & 0.56 & 0.49 & 0.56 & 0.69 & 0.49 \\
& GPT-5mini (3-shot) & 0.56 & 0.46 & 0.60 & 0.49 & 0.46 \\
\midrule
Neural & Feedforward MLP & 0.45 & 0.27 & 0.44 & 0.29 & 0.32 \\
\midrule
Symbolic & AutoAspect$^\dagger$ & 0.39 & 0.23 & 0.40 & --- & --- \\
\midrule
Human & First annotator & \textbf{0.84} & \textbf{0.76} & \textbf{0.84} & \textbf{0.77} & \textbf{0.82} \\
\bottomrule
\end{tabular}
\caption{Baseline results on the test split (254 events, 72 sentences). 
Human performance reflects first-pass annotator accuracy against adjudicated gold labels. 
Precision and Recall are macro averages across classes. 
$^\dagger$AutoAspect results are taken from the original publication and are not directly comparable.}
\label{tab:baselines}
\end{table*}

\subsection{Baseline models}
To establish a performance baseline for this task on this data set, we test on two types of models: 1) open-source and closed LLMs under a simple prompting paradigm; and 2) a simple feedforward neural architecture.

\paragraph{LLM Prompting.}



We experiment with multiple LLMs in a prompting paradigm to evaluate their capability for aspect classification without fine-tuning. LLM performance on structured prediction tasks has been shown to vary drastically based on slight changes to prompt structure~\citep{lu-etal-2022-fantastically}, and other work suggests that LLMs lack meta-linguistic reasoning capability~\citep{bonn-etal-2024-adjudicating}; we examine the ability of LLMs to identify covert aspectual information from a sentence, as well as produce a baseline against which to compare other neural approaches in the future. Although finding the optimal prompt for this task is intractable, we first ran a preliminary search across different prompt strategies on a validation set to determine if any of them boosts aspect prediction performance significantly.\footnote{Details of the prompt-tuning experiments available in Appendix C} We found minimal differences between prompt styles, and proceeded to the test phase. Our tests compare \texttt{Llama-3.1-8b-instruct} \citep{grattafiori2024llama3herdmodels} and \texttt{GPT-5mini}. We experiment with few-shot in-context learning using 3 examples per label (21 examples per prompt). 

\paragraph{Feedforward Classifier.}

We investigate the ability of LLM encoder layers to capture representations that may be useful for aspect classification based on the hypothesis that contextual embeddings encode a broad range of linguistic phenomena~\citep{arora-etal-2024-causalgym}. We do this by combining the token embeddings of the input sentence with a simple feedforward neural classifier to produce a label prediction from the text alone via supervised training.

To evaluate the usefulness of LLM embeddings out-of-the-box, we pass the natural language sentence through the encoder block of Llama-3.1 8B and average the resulting token embeddings to generate a sentence vector with standardized dimensions, then use a simple feedforward classifier head to produce a label prediction. We use the same averaged embedding as input for each event in the sentence. We train a fully-connected feedforward network to predict one of the seven aspect labels using that embedding as input. The results from this method serve as a useful benchmark for evaluating more complex strategies in future work.

\subsection{Automatic Modeling Results}\label{sec:results}


\autoref{tab:baselines} displays the accuracies and F1-scores across the two LLMs and the feedforward neural classifier. We report weighted F1, average precision, and average recall; to address the imbalanced label distribution in the data, we also report macro F1. We evaluate all methods on the same stratified test set. 
The table also shows reported results for AutoAspect \citep{chen2021autoaspect}, a rule-based approach to UMR aspect classification. Note that the reported results are on a different test set, so these serve as a general reference for rule-based approaches rather than a direct comparison. Finally, we show annotation agreement scores as an upper bound for the task.

\paragraph{LLM Prompting.}

The dataset's significant imbalance, with \textit{State} and \textit{NONE} labels being dominant and many classes being rare, heavily influences the results. Performance on minority classes like \textit{Endeavor} is weak regardless of the prompt or model, suggesting that neither prompt engineering nor increased parameter count can overcome severe data sparsity. Weighted F1 scores are skewed by the majority class, while lower macro F1 scores accurately reflect poor performance across most categories. GPT-5mini outperforms Llama across all metrics, which we attribute to architectural updates, including advanced knowledge distillation. GPT-5mini performance is relatively indifferent to in-context examples, while Llama's performance actually decreases with the addition of in-context learning, suggesting that more comprehensive training is needed for aspect prediction.

\paragraph{Feedforward Classifier.}

Neural classification using Llama embeddings results in middling performance, coming short in all three evaluation metrics compared to LLM prompting methods. Although sentence embeddings have been seen to capture semantic information in other tasks, these results demonstrate that embeddings alone are insufficient for capturing aspectual information.

\paragraph{Human Baseline.}

Across the board, our human baseline outperforms all automated aspect annotation methods tested. The human baseline scores represent one annotator's first-pass labels for each of the events in the test set, compared against the final adjudicated labels. The overwhelming success of the human annotator over the automated baselines confirms two things: (i), the complexity of the aspect annotation task, and (ii), the need for automated methods which better utilize the sentential context and/or the inherent graphical nature of event-argument structures.

\section{Conclusion and Future Work}\label{sec:conclusion}

In this work, we introduce a new dataset of English sentences annotated with aspect labels within the UMR framework. We describe the annotation scheme and guidelines used to assign aspectual categories from the UMR aspect lattice and detail our multi-stage annotation and adjudication process.  We establish initial baseline benchmarks using rule-based methods, embedding-based classifiers, and large language model prompting approaches. 
The guidelines, dataset, standard data splits, and initial benchmarks together lay a foundation for studying aspect in structured semantic representations and will support future work on automated UMR parsing and cross-linguistic semantic annotation.

\section*{Ethical Considerations}

This work builds on existing publicly available corpora that were previously released for research purposes. Our dataset adds aspectual annotations to sentences drawn from these sources in accordance with their respective licenses. Annotation was conducted by trained researchers who are authors of this paper.

Because the dataset focuses on English sentences, it representatively only reflects information about English-language corpora and does not directly capture aspectual distinctions present in other languages. Future work will expand this annotation framework to additional languages in order to support broader cross-linguistic semantic analysis.

We do not anticipate significant risks of misuse for this dataset. However, some of the examples in our corpus were pulled from online message fora without censoring, and may contain offensive, explicit, or harmful language. The resource is intended to support research in semantic representation and natural language processing.

\section*{Limitations}

We attempted to reimplement the AutoAspect rules-based classifier \citep{chen-etal-2021-autoaspect} on our novel set of annotated UMR graphs in order to compare its performance against the neural approaches as a benchmark. AutoAspect focuses on a structured set of rules which closely followed the UMR annotation guidelines and decision lattice to predict labels in a wholly deterministic method, without machine learning. However, due to dependency issues with the semantic parser in the original AutoAspect codebase, we are unable to report this benchmark on our dataset, and instead provide the AutoAspect classifier's performance on the dataset with which it was published, as a reference for rule-based approaches in general.

\section{Bibliographical References}\label{sec:reference}

\bibliographystyle{lrec2026-natbib}
\bibliography{custom}

\label{lr:ref}
\bibliographystylelanguageresource{lrec2026-natbib}

\section{Appendix A - Data Statistics and Dataset Splits}\label{sec:appen_a}

\begin{table*}[htbp]
\centering
\begin{tabular}{lcccccccc}
\toprule
Aspect & Little Prince & Minecraft & BOLT DF & WB & Pear Story & Lorelei & UMR 1.0 & Total \\
\midrule
State       & $63$ & $20$ & $119$ & $45$ & $121$ & $0$ & $62$ & $430$ \\
Habitual    & $1$  & $0$  & $17$  & $4$  & $28$  & $0$ & $2$  & $52$ \\
Process     & $2$  & $0$  & $5$   & $5$  & $44$  & $1$ & $1$  & $58$ \\
Activity    & $9$  & $0$  & $31$  & $14$ & $57$  & $0$ & $21$ & $132$ \\
Performance & $35$ & $2$  & $43$  & $18$ & $159$ & $3$ & $57$ & $317$ \\
Endeavor    & $0$  & $0$  & $0$   & $0$  & $2$   & $0$ & $14$ & $16$ \\
\textbf{Total} & $\mathbf{110}$ & $\mathbf{22}$ & $\mathbf{215}$ & $\mathbf{86}$ & $\mathbf{411}$ & $\mathbf{4}$ & $\mathbf{157}$ & $\mathbf{1005}$\\
\bottomrule
\end{tabular}
\caption{Aspect label distribution from existing UMR data before any additional annotation was done.}
\label{tab:gold_label_stats}
\end{table*}

Table~\ref{tab:gold_label_stats} shows general statistics for aspect-labeled UMR data prior to this annotation project. Table~\ref{tab:data_stats} shows distributions and counts at the end of \textsc{Phase 1} annotation, before \textsc{Phase 2} adjudication.

Table~\ref{tab:dataset-splits} provides statistics for our dataset splits. Precise Sentence IDs for the splits will be available on our github (once anonymity is no longer required).
\section{Appendix B - Annotation}\label{sec:appen_b}

\begin{table*}[ht]
\centering
\begin{tabular}{lcccc|c|c}
\toprule
\textbf{Aspect} & \textbf{Little Prince} & \textbf{Minecraft} & \textbf{BOLT DF} & \textbf{WB} & \textbf{Existing Labels} & \textbf{Total}\\
\midrule
State       & $172$ & $14$ & $101$ & $14$ & $430$ & $731$ \\
Habitual    & $41$  & $0$  & $4$   & $0$  & $52$  & $97$  \\
Process     & $31$  & $0$  & $38$  & $8$  & $58$  & $135$ \\
Activity    & $15$  & $2$  & $10$  & $3$  & $132$ & $162$ \\
Performance & $163$ & $43$ & $69$  & $32$ & $317$ & $624$ \\
Endeavor    & $15$  & $0$  & $4$   & $0$  & $16$  & $35$  \\
None        & $158$ & $49$ & $121$ & $77$ &  -    & $405$ \\
\hline
\textbf{Total} & $\mathbf{595}$ & $\mathbf{108}$ & $\mathbf{347}$& $\mathbf{134}$ & $\mathbf{1,005}$ & $\mathbf{2,189}$\\
\hline
\hline
\textbf{Fleiss' Kappa} & $0.78$ & $0.82$ & $0.45$ & $0.40$ & - & -\\
\bottomrule
\end{tabular}
\caption{Label distribution by corpus and annotated aspect. We report Fleiss' Kappa between the two initial annotators and do not include disagreements in the reported total.}
\label{tab:data_stats}
\end{table*}


\begin{table}[!t]
\centering
\begin{tabular}{clc}
\toprule
Category & Metric & Value \\
\midrule
\multirow{7}*{Accuracy} & State             & $0.82$ \\
                        & Habitual          & $0.63$ \\
                        & Activity          & $0.52$ \\
                        & Performance       & $0.80$ \\
                        & Endeavor          & $0.11$ \\
                        & Overall Accuracy  & $0.74$ \\
                        & Perfect Accuracy  & $0.35$ \\
\hline
\multirow{2}*{F1}       & Macro F1          & $0.49$ \\
                        & Weighted Macro F1 & $0.76$ \\
\hline
\multirow{2}*{IAA}      & Fleiss' Kappa     & $0.55$ \\
                        & Gwet's AC1        & $0.66$ \\
\bottomrule
\end{tabular}
\caption{Practice Annotation Results: \textit{Overall Accuracy} is the ratio of the total number of correct annotations over the total number of predicates annotated. \textit{Perfect Accuracy} is the ratio of predicates that were correctly annotated by all annotators. No occurrences of \textit{Process} aspects were present in the practice set.}
\label{tab:practice_anno}
\end{table}

\begin{table}[t]
\centering
\small
\begin{tabular}{lrr}
\toprule
\textbf{Split} & \textbf{Sentences} & \textbf{Events} \\
\midrule
Train & 333 & \phantom{,0}999 \\
Dev   &  71 & \phantom{,0}220 \\
Test  &  72 & \phantom{,0}254 \\
\midrule
\textbf{Total} & \textbf{476} & \textbf{1{,}473} \\
\bottomrule
\end{tabular}
\caption{Stratified 70/15/15 train/dev/test split, divided at the sentence level using dominant aspect label for stratification.}
\label{tab:dataset-splits}
\end{table}

\begin{table}
\centering
\begin{tabular}{lcl} 
\toprule
Gold label & Error rate&Error Label\\
\midrule
Activity & 6.7\% &Performance \\
Endeavor & 12.2\% &Performance \\
Habitual & 6.8\% &Performance \\
None & 6.0\% &State \\
Performance & 3.9\% &None \\
Process & 9.1\% &None \\
State & 4.8\% &None \\
\bottomrule
\end{tabular}
\caption{Final gold labels and the percent for each that was incorrectly annotated then fixed during adjudication out of the total number of annotations for that label. The error label is the most common incorrect annotation.}


\label{tab:annotation_counts}
\end{table}




\begin{figure*}[t]
  \includegraphics[width=\linewidth]{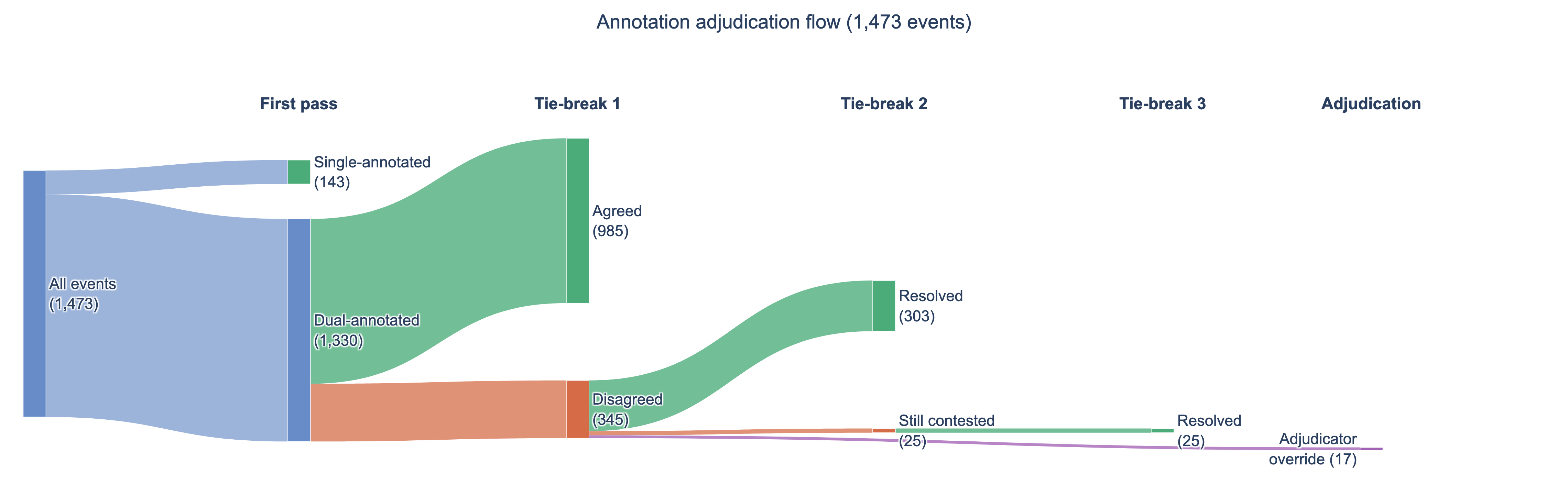}
  \caption{Flow diagram illustrating the annotation and adjudication phases, and how many labels were completed at each stage of the process.}
  \label{fig:annotation-flowchart}
\end{figure*}

Training materials are primarily drawn from existing UMR tutorial materials and supplemented with custom task-specific resources, including
an accessible slide deck
\footnote{\url{https://docs.google.com/presentation/d/1QUAnh2LWlgfvp_0NAj7K-YEdAbjG7zXmCLfEq0sfoa0/edit?usp=sharing}}
which summarizes the UMR guidelines \footnote{\url{https://github.com/umr4nlp/umr-guidelines/blob/master/guidelines.md}} with added clarifications and examples. 

Table \ref{tab:practice_anno} shows the results from the Pear Story practice annotation task. 
Due to the different number of annotations each person performed, we report Gwet's AC1 as a measure for inter-annotator agreement (IAA) since this metric can be calculated for different numbers of labels. We report Fleiss' Kappa only for predicates that were labeled by all annotators. We find moderate-to-good IAA for the practice round, motivating the need for additional training that was conducted.

\autoref{fig:annotation-flowchart} illustrates the flow of data through the two phases of corpus building, including multiple rounds of tie-breaking. The 143 events listed as single-annotated in the first-pass were part of a teaching demonstration, but they did ultimately receive second annotations and were reviewed for consistency during adjudication; this detail was omitted from the diagram for visual clarity.

\section{Appendix C - Modeling}\label{sec:appen_c} 

In this Appendix we provide additional information on the automatic aspect modeling design and results. 

\paragraph{LLM Prompt Tuning.} We try three strategies to gauge the impact of prompt structure on LLM performance: Initially, we manually draft a list of short definitions for each aspect class based on the experience gained from our annotator training sessions. In a second prompt attempt, we provide the initial prompt and instruct the model to generate a better prompt for our task, to investigate if the LLM's pretraining contains aspectual knowledge beyond our basic definitions, which resulted in a streamlined version with more general task instruction. Finally, to take advantage of extensive LLM context windows, we try providing the UMR guidelines for aspect \footnote{\href{github.com/umr4nlp/umr-guidelines/blob/master/guidelines.md\#part-3-3-1-Aspect}{UMR Guidelines GitHub Repository}} in their entirety and instructing the model to predict a label. We find marginally higher validation accuracy with strategy 2, and employ it in testing.

\end{document}